# SymptomWise: A Deterministic Reasoning Layer for Reliable and Efficient AI Systems

**Authors:** Isaac Henry[1], Avery Byrne[1], Christopher Giza[2], Ron Henry[3], Shahram Yazdani[2]

1. Symptomwise.org
2. University of California Los Angeles, David Geffen School of Medicine
3. University of Southern California

## Disclaimer

This document describes the system-level architecture and operational principles of the SymptomWise framework. Certain implementation details, parameterizations, and optimization strategies are not disclosed.

## Abstract

AI-driven symptom analysis systems face persistent challenges in reliability, interpretability, and hallucination. End-to-end generative approaches often lack traceability and may produce unsupported or inconsistent diagnostic outputs in safety-critical settings.

We present SymptomWise, a framework that separates language understanding from diagnostic reasoning. The system combines expert-curated medical knowledge, deterministic codex-driven inference, and constrained use of large language models. Free-text input is mapped to validated symptom representations, then evaluated by a deterministic reasoning module operating over a finite hypothesis space to produce a ranked differential diagnosis. Language models are used only for symptom extraction and optional explanation, not for diagnostic inference.

This architecture improves traceability, reduces unsupported conclusions, and enables modular evaluation of system components. Preliminary evaluation on 42 expert-authored challenging pediatric neurology cases shows meaningful overlap with clinician consensus, with the correct diagnosis appearing in the top five differentials in 88% of cases. Beyond medicine, the framework generalizes to other abductive reasoning domains and may serve as a deterministic structuring and routing layer for foundation models, improving precision and potentially reducing unnecessary computational overhead in bounded tasks.

# 1. Introduction and Motivation

AI systems capable of interpreting patient-reported symptoms have the potential to improve access to medical information and support earlier clinical engagement.[1,2] However, existing approaches frequently rely on opaque or end-to-end generative models that blur the boundary between language understanding and medical reasoning.[3,4] This conflation introduces risks related to hallucination, inconsistency and limited auditability.

In healthcare contexts, reliability and interpretability are not optional.[1,2] Systems must be capable of explaining how conclusions are reached, constrain outputs to known medical knowledge and support expert oversight. SymptomWise is motivated by the need for AI systems that respect these constraints while still leveraging the strengths of modern language models. The methodology proposed here is that reliable AI in safety-critical domains requires separating language understanding from structured abductive reasoning within bounded hypothesis spaces.

# 2. Design Goals and Principles

The SymptomWise platform is guided by the following design goals:

- **Determinism and traceability:** Diagnostic outputs must be reproducible and traceable to explicit symptom inputs and curated knowledge.
- **Human comprehensibility:** Core medical knowledge should be represented in a form that can be reviewed and validated by domain experts.
- **Constrained generative AI:** Language models should augment, not replace, structured reasoning.
- **Modularity:** System components should be independently testable and improvable.
- **Safety-first operation:** The system should minimize hallucination risk and support clinical oversight.

These principles inform all architectural decisions described in this document.

# 3. System Overview

SymptomWise is a hybrid AI platform that does reliable diagnosis by combining curated medical knowledge, deterministic reasoning and constrained use of LLMs. It avoids end-to-end generative diagnosis, instead decomposing the diagnostic process into a sequence of steps, each with clearly bounded responsibilities and evaluable outputs.

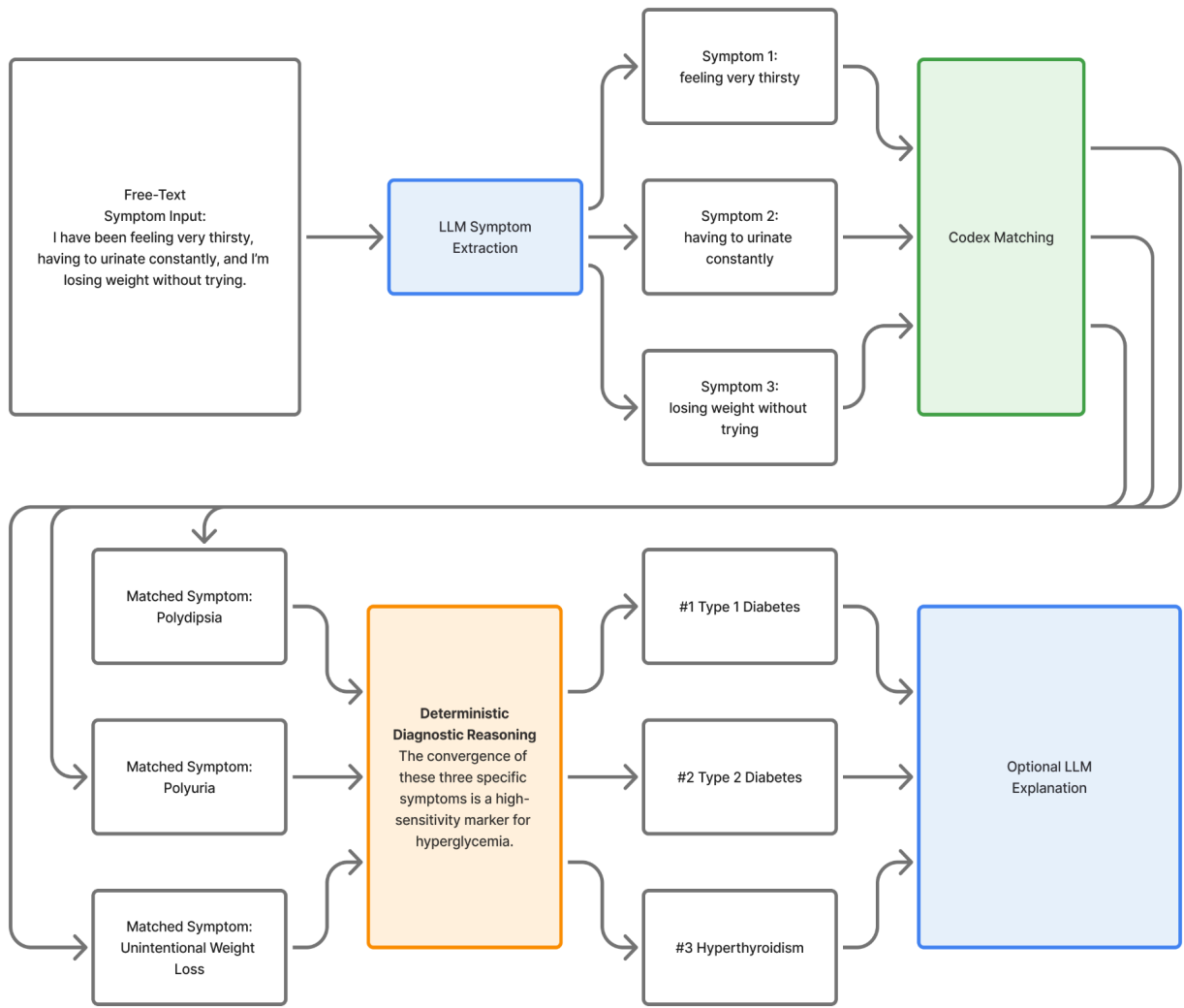

Figure 1. The platform decomposes symptom analysis into auditable stages. Free-text user input is interpreted by a language model solely for symptom extraction. Extracted concepts are semantically normalized and matched to an expert-curated codex. A proprietary deterministic reasoning module operates exclusively over validated codex symptoms to produce a ranked differential diagnosis. Language models may optionally generate user-facing explanations but do not participate in diagnostic reasoning.

First, unstructured text is converted into a set of known symptoms from a curated knowledge base. Then a proprietary deterministic reasoning module is used to generate a ranked differential diagnosis. Finally the diagnosis can be, optionally, fed into generative models to produce user-friendly explanations constrained to those results.

## 3.1 Architectural Principles

The architecture is guided by the following principles:

- Diagnostic authority resides in deterministic reasoning, not generative models.
- Language models are limited to bounded roles, primarily input interpretation and optional explanation.
- Knowledge is explicit, finite, and auditable through an expert-curated codex.

- Reasoning is constrained to the smallest adequate hypothesis space, improving interpretability and limiting unsupported generative drift.
- System interfaces are stable and inspectable, so each stage produces bounded outputs for the next.
- Modularity supports safety and improvement, since extraction, normalization, reasoning, and explanation can be independently evaluated.

Together, these principles improve clinical reliability while also suggesting a broader AI design pattern: deterministic structuring layers can increase precision and potentially reduce the computational inefficiency of reasoning over unnecessarily broad conceptual spaces.

## 3.2 Knowledge Foundation

The core of the system is a structured codex of symptoms and diagnoses developed and maintained by human domain experts, with AI-assisted tooling used to support coverage, consistency and maintainability. Each diagnosis in the codex is associated with a defined set of relevant symptoms, forming an explicit representation of known medical relationships.

The codex serves as the authoritative reference for diagnostic reasoning. It is versioned and auditable, enabling expert review and controlled evolution over time. Specific internal schemas and validation mechanisms are intentionally abstracted.

## 3.3 Symptom Interpretation and Normalization

User input consists of free-text descriptions of symptoms. A language model is used strictly for information extraction, identifying symptom concepts present in the text and producing structured representations.

Each codex symptom is represented in a shared semantic embedding space. Extracted symptom representations are mapped into this space and matched against codex-defined symptoms based on semantic proximity. It needs to be noted that the embedding layer is a mapping mechanism, not a reasoning mechanism. This process enables robust handling of linguistic variability while ensuring that downstream reasoning operates exclusively on validated symptom definitions.

## 3.4 Deterministic Diagnostic Reasoning

The presence or confirmed absence of validated codex-matched symptoms is represented as a binary observation vector of length N, where N is the number of known symptoms in the codex. This vector is then supplied to the proprietary diagnostic reasoning module, which generates a ranked differential diagnosis over the finite set of codex-defined diseases.

The reasoning layer is deterministic and reproducible, and it does not rely on generative inference. Rather than allowing a language model to propose diagnostic hypotheses, the module evaluates the degree to which the observed symptom configuration evokes each candidate disease within the bounded codex. In this framework, symptom–disease relationships are not treated as uniform. A given symptom may carry different explanatory significance for different diseases. Accordingly, the model assigns disease-conditioned evoking values to symptom–disease pairings using a proprietary mathematical framework derived in part from interaction-based formalisms used in the physical sciences. These evoking values quantify the extent to which an observed symptom supports, weakens, or fails to support a given diagnostic hypothesis.

At a conceptual level, the reasoning module aggregates these disease-specific evoking relationships across the full observed symptom set to evaluate explanatory coherence between the patient presentation and each codex-defined disease signature. This evaluation incorporates three broad forms of structural evidence: the presence of expected supportive features, the absence of expected features, and the presence of features that are poorly aligned or in tension with the candidate disease. The output is an ordered, weighted differential diagnosis over the permissible hypothesis space.

The exact functional form, parameterization, and optimization strategy of this reasoning module constitute proprietary intellectual property and are not disclosed in this document. However, several properties are central to the architecture. First, inference is bounded to a finite set of expert-curated diagnostic hypotheses. Second, the same validated symptom input always yields the same ranked output. Third, each result is traceable to explicit codex-defined symptom relationships rather than opaque end-to-end generation. This structure supports auditability, expert review, and modular evaluation in safety-critical diagnostic settings.

## 3.5 Optional Generative Output

In optional downstream steps, the ranked differential diagnosis may be supplied to a language model to generate user-facing explanations or summaries. In this role, the language model is constrained to the diagnoses produced by the deterministic reasoning layer and does not introduce new diagnostic hypotheses.

# 4. Generalized Codex-Driven Reasoning Framework

At the core of the system is a codex-driven reasoning framework that is not specific to medicine, but instead represents a general class of abductive inference problems. This section formalizes the framework and characterizes the types of problems it can address.

## 4.1 Codex Representation

Let $H = \{h_1, \dots, h_n\}$ denote a finite set of hypotheses (e.g., diagnoses, failure modes, root causes), and let $F = \{f_1, \dots, f_m\}$ denote a finite set of observable features. The codex defines a **binary incidence relation** between hypotheses and features. For each hypothesis $h \in H$, the codex specifies a feature set:

$$C(h) \subseteq F$$

indicating which features are associated with that hypothesis. Equivalently, the codex may be represented as a binary vector:

$$c(h) \in \{0, 1\}^m$$

where each dimension corresponds to a feature in $F$ and membership is strictly binary.

No weights or probabilistic parameters are required at the codex level; all relationships are defined by expert-validated inclusion or exclusion.

## 4.2 Observations Input Set

An input case is represented as a validated binary observation vector:

$$x \in \{0, 1\}^m$$

where $x_i = 1$ indicates the confirmed presence of a feature $f_i$, and $x_i = 0$ indicates its absence.

## 4.3 Hypotheses Output Set

The output is a weighted ordered set of hypotheses (diagnoses) in the codex. The weighting does not have physical meaning but is taken as a confidence factor for each hypothesis.

## 4.4 Class of Solvable Problems

The codex-driven framework applies to a broad class of abductive reasoning problems characterized by bounded explanatory spaces and structured evidence. These problems share several defining properties:

**Finite hypothesis space**

The set of possible explanations is bounded, limited, enumerable, and semantically stable. Hypotheses are not generated ad hoc but drawn from a curated and versioned library of permissible candidates.

**Discrete, validated features**

Observations can be mapped to a defined feature set in which both presence and confirmed absence are meaningful. The evidentiary space is structured rather than open-ended.

**Many-to-many relationships**

Features are shared across multiple hypotheses, and individual hypotheses may manifest through multiple features. Ambiguity is inherent and must be resolved through structured comparison rather than single-feature triggers.

**Explanatory objective**

The goal is to identify which hypothesis (or ranked set of hypotheses) most coherently explains the observed evidence. The framework is abductive rather than predictive: it seeks explanatory alignment, not probabilistic forecasting of future states.

When these conditions are satisfied, inference can be formulated as structured matching between observed feature configurations and curated hypothesis signatures. The resulting ranked outputs preserve interpretability while acknowledging ambiguity. Rather than asserting calibrated probabilistic certainty, the system evaluates explanatory coherence within a bounded, expert-defined space.

This structure is not specific to medicine. It applies wherever explanations are finite, evidence is structured, and disciplined ranking is preferable to unconstrained generation. The subsequent sections illustrate how this general reasoning form extends across diagnostic, causal, regulatory, and optimization domains.

## 4.5 Relationship to Probabilistic Models

While structurally related to binary probabilistic models (e.g., Bernoulli Naive Bayes), the SymptomWise framework does not require calibrated probabilities or full generative assumptions.[5] Rather than estimating likelihoods over an unconstrained outcome space, it evaluates explanatory coherence within a finite, expert-defined hypothesis set. Probabilistic methods aim at calibrated prediction under uncertainty; the codex-driven framework instead prioritizes structured alignment between observed evidence and permissible explanations. The resulting ranked outputs remain interpretable without asserting probabilistic certainty.

## 4.6 Disease-Conditioned Evoking Structure

A central property of the reasoning framework is that observable features are not assumed to contribute uniformly across hypotheses. The same feature may be weakly informative for one disease, highly characteristic of another, and potentially contradictory for a third. The system therefore treats symptom–disease relationships as disease-conditioned rather than globally fixed.

This disease-conditioned notion of evocation is especially important in medicine, where common symptoms such as headache, fatigue, weakness, or vomiting may have very different explanatory significance depending on the surrounding syndrome. By preserving this asymmetry explicitly within the reasoning layer, the framework can support more disciplined differential ranking while remaining fully bounded by expert-curated diagnostic knowledge.

# 5. Reliability and Hallucination Mitigation

Our proposed system reduces hallucination risk through architectural separation and constraint:

- Diagnoses are selected from a finite, expert-curated set
- Language models are restricted to non-authoritative roles
- Diagnostic reasoning is deterministic and traceable
- Each processing stage can be independently evaluated

By grounding all diagnostic outputs in curated knowledge and deterministic reasoning, the system avoids spurious conclusions commonly associated with unconstrained generative approaches.[3,4]

# 6. Evaluation and Modularity

The modular structure of SymptomWise enables targeted evaluation at each stage of the pipeline rather than requiring the system to be assessed only as a monolithic black box. This decomposition is important in safety-critical settings because different failure modes arise at different stages of processing.

At the language interpretation stage, symptom extraction can be evaluated using standard information extraction metrics such as precision, recall, and agreement with expert-annotated symptom concepts. At the normalization stage, semantic matching can be assessed in terms of mapping quality, consistency, and robustness to linguistic variation. At the reasoning stage, diagnostic performance can be evaluated against expert reference standards using ranking-based measures such as Top-k inclusion, rank position of reference diagnoses, and structured expert review of the plausibility of alternative hypotheses.

This separation provides two practical advantages. First, it allows individual components to be improved without destabilizing the full system. Second, it enables more precise safety analysis by making it possible to distinguish failures of extraction, normalization, codex coverage, and deterministic inference. Such modular evaluation is particularly important in clinical applications, where transparency about the source of errors is often as important as aggregate performance.

## 6.1 Preliminary Pediatric Neurology Evaluation

As an initial validation of the deterministic codex-driven framework in a high-stakes domain, SymptomWise was evaluated on 42 expert-authored pediatric neurology clinical vignettes. Each case was independently reviewed by SymptomWise and by four practicing pediatric neurologists, whose assessments were used to establish a consensus diagnosis as the reference standard.

Performance was measured by inclusion and ranking of the consensus diagnosis within the system's differential. The system included the consensus diagnosis as the top-ranked result in 30 of 42 cases (71%; 95% CI, 55%–84%), within the top three in 34 cases (81%; 95% CI, 66%–91%), and within the top five in 37 cases (88%; 95% CI, 74%–96%). These results indicate meaningful alignment between the deterministic codex-based framework and expert clinical reasoning in this initial cohort.

Qualitative review suggests that in clinically nonspecific presentations, SymptomWise tends to generate broader differentials, while subspecialists often produce narrower, domain-focused lists. As a result, inclusion-based metrics may underrepresent usefulness in cases where the system identifies a wider but still clinically plausible hypothesis space. In at least one instance, a normal physiologic pattern was classified as pathologic, reflecting the current system's focus on pathologic differentials and underscoring the importance of clinician oversight.

This evaluation has several limitations. The dataset was modest in size and based on expert-authored vignettes rather than prospective clinical data, and the reference standard was clinician consensus rather than outcome-confirmed diagnosis. Additionally, performance varied across subspecialty categories, likely reflecting differences in case structure, ambiguity, and current codex representation. Accordingly, these findings should be interpreted as preliminary validation of the framework rather than a definitive assessment of clinical performance.

These results, accepted for presentation at an international pediatric neurology conference, are included to illustrate that deterministic, codex-driven reasoning can produce differential diagnoses consistent with expert judgment, while preserving traceability and modular auditability.

# 7. Generalization Beyond Medicine

While SymptomWise algorithm is initially applied to medical symptom analysis, the underlying framework applies to a broader class of decision-support problems characterized by:

- A bounded set of explanatory hypotheses
- Observable but incomplete or noisy evidence
- Many-to-many relationships between evidence and explanations
- A requirement for ranked outputs and interpretability

Examples include fault diagnosis, root-cause analysis, compliance triage, and other expert reasoning domains where structured knowledge and deterministic inference are preferred over unconstrained prediction.

## 7.1 Diagnosis of System States Under Ambiguity

In engineered environments, diagnostic reasoning typically concerns the identification of a system's present state within a bounded set of physically or logically possible configurations. These states may include conditions such as sensor malfunction, valve failure, thermal overload, or degraded seal integrity. The hypothesis space is finite and defined by the architecture of the system itself.

The difficulty arises not from the number of possible states alone, but from ambiguity in the observable evidence. Sensor readings may be noisy, delayed, or internally inconsistent. A single abnormal measurement, such as elevated temperature, may be compatible with several distinct failure states. Conversely, one primary fault may generate a cascade of secondary indicators across different components.

Within this setting, the model evaluates the total configuration of observed features against a structured library of known system states. Each state is defined through expert-specified relationships to observable indicators. Rather than treating signals independently, the framework assesses how well each candidate state accounts for the pattern as a whole. This allows it to distinguish between signals that are central to a fault and those that are merely downstream effects.

The output is a ranked list of plausible system states. This ranking provides a disciplined starting point for investigation: engineers can begin with the explanation that most coherently accounts for the full pattern of evidence. By operating within a finite and explicitly defined state space, the framework avoids speculative diagnoses and supports efficient, technically defensible troubleshooting in complex mechanical and digital systems.

## 7.2 Abductive Reconstruction of Causal Origins

Root-cause analysis differs in emphasis from state diagnosis. While state diagnosis asks what condition best explains a present configuration of signals, root-cause analysis asks why a sequence of events occurred. The object of reasoning is not merely the system's current state, but the causal origin of a failure within a structured network of dependencies.

In domains such as software infrastructure, manufacturing processes, or large-scale IT systems, events propagate through defined pathways. A memory leak may lead to service instability, which in turn triggers cascading outages across dependent components. Alternatively, multiple conditions, such as high traffic combined with insufficient resource allocation, may converge to produce a single visible failure. The relationships are many-to-many and frequently non-linear.

Here, the model again operates within a bounded hypothesis space: a finite set of failure modes or initiating causes that are known to be possible within the architecture. Observed events are compared against these candidate causes, each defined by expert-informed relationships to downstream manifestations. The reasoning process evaluates which underlying cause most coherently accounts for the total pattern of observed consequences.

Unlike simple event correlation, this approach preserves logical discipline. It does not search indefinitely for speculative explanations, but restricts inference to structurally permissible origins. The resulting ranked output identifies the most plausible initiating causes while preserving alternative explanations for review.

Root-cause analysis carries implications beyond immediate repair. It informs accountability, long-term remediation, and structural redesign. For this reason, transparency and interpretability are essential. By grounding inference in a finite, expert-defined causal framework and producing a ranked, explainable set of candidates, the model supports durable corrective action rather than temporary symptom suppression.

## 7.3 Normative Classification Under Regulatory and Legal Constraint

Compliance triage differs from both state diagnosis and root-cause reconstruction. The central question is not simply w*hat condition exists or what causes a sequence of events*, but whether a set of actions or patterns satisfies, or violates a defined normative framework. The reasoning task is classificatory rather than predictive: observed behavior must be mapped onto a bounded set of legally or regulatorily defined categories.

In healthcare, this often arises under statutory regimes such as HIPAA's Privacy and Security Rules. The hypothesis space is constrained to legally recognized violations or compliance states. For example, impermissible disclosure of protected health information, unauthorized access ("snooping"), inadequate safeguards, or compliant access consistent with treatment, payment, or operations. The evidence may consist of electronic health record access logs, billing data, device telemetry, or audit trails. Individual data points may appear benign in isolation, yet in combination they may form a pattern indicative of misconduct.

The challenge is compounded by intentional obfuscation. A single access event may be justifiable, but repeated access across unrelated patient charts, or access outside normal work patterns, may shift the interpretation. Compliance reasoning therefore requires evaluating how multiple features jointly support or undermine specific regulatory classifications.

The same structural problem appears outside medicine. In contractual compliance, for example, a set of actions must be evaluated against defined obligations: performance milestones, confidentiality clauses, non-compete provisions, or disclosure requirements. A missed deadline alone may not constitute breach; however, a pattern of delayed deliverables, unapproved subcontracting, and inconsistent reporting may collectively support classification as material non-compliance. The hypothesis space is again bounded, not by mechanical states, but by legally recognized categories such as breach, substantial performance, force majeure exception, or compliant execution.

Similarly, in legal advocacy contexts, such as tools supporting patients, tenants, or clients, the task may involve classifying documented events into statutory rights violations. For instance, housing advocacy platforms must determine whether a pattern of landlord behavior constitutes unlawful eviction, discriminatory practice, or permissible action under local housing codes. The reasoning must align with explicit legal definitions rather than abstract risk scores.

The framework operates by mapping validated features to expert-defined legal or regulatory categories and producing a ranked set of plausible classifications. Importantly, it does not invent new violations or speculate beyond codified definitions. Instead, it evaluates how well the observed evidence aligns with each permissible classification within the governing framework.

This ranking is not merely operationally useful; it is institutionally necessary. Regulatory enforcement, contractual disputes, and rights-based advocacy carry material consequences. Decisions must be explainable to oversight bodies, courts, regulators, or opposing counsel. By grounding inference in explicit, curated relationships and producing traceable outputs, the model replaces opaque risk scoring with structured, auditable reasoning.

In this way, compliance triage represents a third application of the codex-driven framework: not the diagnosis of system states, nor the reconstruction of causal origins, but the disciplined classification of behavior under binding normative constraints.

# 8. Future Directions

Ongoing work includes improving symptom extraction accuracy, refining semantic matching by modeling symptoms as regions in embedding space, expanding the codex, and applying the framework to additional domains that benefit from structured knowledge and deterministic reasoning.

## 8.1 Conceptual Fencing and Controlled Model Boundaries

Another prospective extension of the codex-driven framework involves establishing structured "fences" around sensitive or prohibited conceptual domains. In contrast to treatment optimization, where the objective is to identify the best-aligned protocol, fencing aims to detect when model reasoning approaches predefined high-risk conceptual regions and to intervene in a principled manner.

Many institutions require language models to avoid or carefully manage topics such as bioterrorism, online child exploitation, suicide facilitation, or other forms of high-risk content. The difficulty lies in distinguishing between superficially similar queries that differ materially in intent; for example, a legitimate chemistry homework question versus instructions for weaponization, or a clinical discussion of depression versus encouragement of self-harm.

This problem can be formulated analogously to diagnosis. Instead of hypotheses representing diseases or treatment protocols, they represent bounded conceptual risk categories. Each category can be encoded as a structured feature set derived from expert curation, including semantic indicators, contextual markers, and domain-specific patterns that meaningfully characterize the concept. User input, model tokens, or intermediate reasoning states can be mapped into the same representational space.

The task then becomes one of structured proximity assessment: determining whether a given input or reasoning trajectory sufficiently aligns with a defined risk category to warrant intervention. Rather than relying on opaque probability thresholds or ad hoc keyword filters, the system evaluates whether the feature configuration meaningfully matches a curated conceptual signature.

Intervention strategies can be graduated. A high-confidence match may trigger refusal and predefined escalation workflows. Lower-confidence proximity may prompt clarification, redirection, or constrained responses. Because the categories are explicitly defined and versioned, the resulting behavior is auditable and adjustable over time.

In this formulation, fencing is not merely reactive content moderation. It is the disciplined classification of conceptual territory within a bounded hypothesis space. By applying the same deterministic, codex-driven reasoning structure used in diagnosis and treatment optimization, the framework offers a transparent method for controlling model boundaries while minimizing both over-blocking and under-detection.

## 8.2 Protocol Optimization and Precision Therapeutics

While SymptomWise is initially focused on diagnostic reasoning, the same codex-driven framework extends naturally to therapeutic optimization. In oncology and other complex chronic conditions, treatment selection involves choosing among structured, multi-agent protocols rather than prescribing a single medication. The objective is not merely to select a permissible therapy, but to identify the protocol most likely to produce the best outcome for a specific patient.

Modern regimens often include combinations of medications administered at defined dosages, intervals, and durations. Selection depends on staging, biomarkers, prior treatment exposure, comorbidities, and evolving standards of care. With the rise of precision medicine, additional variables now materially influence outcomes:

pharmacogenomic variants affecting drug metabolism (e.g., CYP450 polymorphisms), receptor or ion channel mutations altering therapeutic responsiveness, tumor-specific molecular markers guiding targeted agents, and clinically significant drug–drug interactions.[6,7]

This problem can be formulated analogously to diagnosis. Instead of hypotheses representing diseases, they represent candidate treatment protocols. Each protocol may be encoded as a structured feature set describing its indications, biomarker requirements, contraindications, toxicity profiles, interaction constraints, and dose-adjustment rules. A patient's clinical and molecular profile likewise forms a validated feature representation incorporating disease subtype, staging, laboratory data, genomic information, comorbidities, and concurrent medications.

The reasoning task becomes one of structured alignment: determining which treatment protocol most coherently matches the patient's full profile under established medical constraints. Rather than producing a generalized recommendation based on population averages, the framework evaluates how well each candidate protocol "fits" the individual patient configuration. The output is a ranked set of protocols, prioritizing those whose structured characteristics align most closely with the patient's biological and clinical context.

In this way, the same mathematical structure used to identify the most plausible diagnosis can be extended to identify the most appropriate therapeutic strategy. The system does not invent novel treatments or override clinical standards; instead, it organizes and evaluates established protocols against individualized patient data in a transparent, auditable manner.

## 8.3 Deterministic Structuring, Routing, and Anchoring for Foundation Models

A broader implication of the SymptomWise framework is that deterministic codex-driven reasoning may serve as a structuring, routing, and anchoring layer for foundation models.[8,9] Rather than replacing large language models, this approach enhances them by transforming free-form input into validated concepts, constraining reasoning to a finite expert-defined hypothesis space, and identifying the subset of pathways most worthy of downstream generative attention. In bounded or semi-bounded domains, this may improve accuracy by reducing unsupported hypothesis drift, improve interpretability by making the active reasoning space explicit and auditable, and improve efficiency by reducing unnecessary generative branching and repeated exploration of implausible pathways.[8,9] In this sense, the deterministic layer functions as a front-end orchestration mechanism: it sharpens input conditioning, stabilizes output generation, and may reduce token usage, latency, compute burden, and associated energy costs in high-volume or high-stakes settings.[10,11] This should be understood as a testable engineering hypothesis rather than a settled claim, and future work should evaluate such architectures directly using matched benchmarks for task accuracy, token consumption, latency, compute utilization, and energy use.

## 9. Conclusion

SymptomWise presents a structured alternative to end-to-end generative diagnosis through a hybrid architecture grounded in bounded hypothesis spaces, expert-curated knowledge, and deterministic abductive reasoning. Rather than treating language models as autonomous decision-makers, the framework separates language understanding from diagnostic authority, placing generative capabilities within an explicit, auditable reasoning structure.

At its core, SymptomWise models clinical inference as structured matching between validated feature configurations and curated hypothesis signatures. This supports ranked differentials, preserves interpretability under ambiguity, and enables traceability from outputs to underlying evidence. By emphasizing explanatory coherence within a finite hypothesis space, the system favors disciplined alignment over unconstrained generation. Preliminary pediatric neurology results further suggest meaningful overlap with specialist clinician consensus while preserving traceability and modular auditability.

The broader contribution extends beyond medical symptom analysis. The same codex-driven framework applies to other abductive reasoning domains, including state diagnosis, causal reconstruction, regulatory classification, therapeutic optimization, conceptual boundary control, and stabilization of generative AI systems. Across these settings, the core principle is consistent: reliable decision support in safety-critical environments requires bounded explanatory spaces, explicit knowledge representation, and deterministic structure.

For academic collaborators, this work offers both a practical architecture and a research direction. Future study may examine the theoretical properties of bounded abductive inference, empirical validation across clinical cohorts, comparative analysis with probabilistic models, and integration strategies for hybrid generative-deterministic systems. The aim is not to replace statistical learning, but to complement it with logical scaffolding that improves robustness, interpretability, and institutional trust.

More broadly, the framework suggests a systems role for advanced AI. In domains where plausible explanations, classifications, or actions can be explicitly bounded, deterministic codex-driven reasoning may serve as a front-end orchestration layer for foundation models, reducing unnecessary exploratory computation while improving precision, interpretability, and stability. This raises the possibility that reliability-oriented architectures may also support more compute-efficient and environmentally responsible AI deployment.

In an era when AI systems increasingly influence high-stakes decisions, SymptomWise proposes a disciplined path forward: generative intelligence constrained, stabilized, and sharpened through expert-aligned structure.

This preserves your meaning, removes some recurrence, and avoids slightly inflated phrasing like "potentially made more efficient" in the final sentence, which is already covered more carefully in the preceding paragraph.

# Reference:

**1.** Miller, R. A. (1994). *Medical diagnostic decision support systems—past, present, and future: A threaded bibliography and brief commentary.* Journal of the American Medical Informatics Association, 1(1), 8–27.

**2.** Shortliffe, E. H. (1976). *Computer-Based Medical Consultations: MYCIN.* New York: Elsevier.

**3.** Ji, Z., Lee, N., Frieske, R., Yu, T., Su, D., Xu, Y., Ishii, E., Bang, Y., Chen, D., Dai, W., Chan, H. S., Madotto, A., & Fung, P. (2023). *Survey of hallucination in natural language generation.* ACM Computing Surveys, 55(12), Article 248.

**4.** Bender, E. M., Gebru, T., McMillan-Major, A., & Shmitchell, S. (2021). *On the dangers of stochastic parrots: Can language models be too big?* In Proceedings of the 2021 ACM Conference on Fairness, Accountability, and Transparency (pp. 610–623).

**5.** McCallum, A., & Nigam, K. (1998). *A comparison of event models for Naive Bayes text classification.* In Proceedings of the AAAI-98 Workshop on Learning for Text Categorization.

**6.** Relling, M. V., & Evans, W. E. (2015). *Pharmacogenomics in the clinic.* Nature, 526(7573) 343–350.

**7.** Garraway, L. A., & Lander, E. S. (2013). *Lessons from the cancer genome.* Cell, 153(1), 17–37.

**8.** Shazeer, N., Mirhoseini, A., Maziarz, K., Davis, A., Le, Q., Hinton, G., & Dean, J. *Outrageously large neural networks: The sparsely-gated mixture-of-experts layer.* International Conference on Learning Representations(2017).

**9.** Fedus, W., Zoph, B., & Shazeer, N. (2022). *Switch Transformers: Scaling to trillion parameter models with simple and efficient sparsity.* Journal of Machine Learning Research, 23(120), 1–39.

**10.** Patterson, D., Gonzalez, J., Le, Q. V., Liang, C., Munguia, L.-M., Rothchild, D., So, D., Texier, M., & Dean, J. (2021). *Carbon emissions and large neural network training.* CoRR, abs/2104.10350.

**11.** Wu, C.-J., Raghavendra, R., Gupta, U., Acun, B., Ardalani, N., Maeng, K., Chang, G., Aga Behram, F., Huang, J., Bai, C., et al. (2022). *Sustainable AI: Environmental implications, challenges and opportunities.* Proceedings of Machine Learning and Systems, 4, 795–813.